\documentclass[11pt]{article}
\usepackage[top=0.6in,left=1.2in,right=1.2in,footskip=0.75in]{geometry}

\usepackage{changepage}

\usepackage[utf8]{inputenc}

\usepackage{textcomp,marvosym}

\usepackage{fixltx2e}

\usepackage{amsmath,amssymb}

\usepackage{cite}

\usepackage{nameref,hyperref}

\usepackage[right]{lineno}

\usepackage{microtype}
\DisableLigatures[f]{encoding = *, family = * }

\usepackage{rotating}
\usepackage{multirow}
\usepackage{CJKutf8}
\usepackage{mathtools}
\usepackage{tabu}
\usepackage[section]{placeins}

\date{}

\begin{document}
\vspace*{0.35in}

\begin{flushleft}
{\Large
\textbf\newline{Language Detection For Short Text Messages In Social Media}
}
\newline
\\
Ivana Balažević\textsuperscript{1},
Mikio Braun\textsuperscript{1},
Klaus-Robert Müller\textsuperscript{1, 2}
\\
\bigskip
\bf{1} Machine Learning Group, Technische Universität Berlin, Berlin, Germany
\\
\bf{2} Department of Brain and Cognitive Engineering, Korea University, \\Seoul, Korea
\\
\bigskip







\end{flushleft}
\begin{center}
\section*{Abstract}
\end{center}

\begin{changemargin}{0.3in}{0.3in}
With the constant growth of the World Wide Web and the number of documents in different languages accordingly, the need for reliable language detection tools has increased as well. Platforms such as Twitter with predominantly short texts are becoming important information resources, which additionally imposes the need for short texts language detection algorithms.
In this paper, we show how incorporating personalized user-specific information into the language detection algorithm leads to an important improvement of detection results. To choose the best algorithm for language detection for short text messages, we investigate several machine learning approaches. These approaches include the use of the well-known classifiers such as SVM and logistic regression, a dictionary based approach, and a probabilistic model based on modified Kneser-Ney smoothing. Furthermore, the extension of the probabilistic model to include additional user-specific information such as evidence accumulation per user and user interface language is explored, with the goal of improving the classification performance.
The proposed approaches are evaluated on randomly collected Twitter data containing Latin as well as non-Latin alphabet languages and the quality of the obtained results is compared, followed by the selection of the best performing algorithm. This algorithm is then evaluated against two already existing general language detection tools: \texttt{Chromium Compact Language Detector 2 (CLD2)} and \texttt{langid}, where our method significantly outperforms the results achieved by both of the mentioned methods.
Additionally, a preview of benefits and possible applications of having a reliable language detection algorithm is given.
\end{changemargin}
\restoregeometry

\section{Introduction}
Language detection is a natural language processing task of identifying the language a given document is written in. It is often the first step in a document processing pipeline. Moreover, it is considered to be a critical preprocessing step in applications that require language-specific modeling, such as search engines, where depending on the detected language different tokenizers may be used.  Another common example of applying language detection is as a preceding step to machine translation, since the language of the text to be translated is not always specified. Therefore, a reliable language detection tool is needed. 

Even though language detection itself has been studied since the 1960s, short texts appearing in social media websites and forums still require a special treatment, due to their specific language type, i.e. acronyms, abbreviations, spelling mistakes, emoticons, new words, etc. Therefore, despite the fact that language detection has been a long-known problem, an appropriate solution for short texts classification is yet to be found. Platforms such as Twitter, where these short texts are used, are recently becoming important real-time information resources \cite{Carter:2013:MLI:2447287.2447296, Golovchinsky10makingsense}. A wide range of applications is connected to their usage - event detection \cite{Sakaki:2010:EST:1772690.1772777, Vieweg:2010:MDT:1753326.1753486}, media analysis \cite{altheide}, opinion mining \cite{Jansen:2009:TPT:1656289.1656297, tumasjan2010predicting}, predicting movie ratings \cite{Oghina:2012:PIM:2260641.2260703}, etc. The majority of the social networks users contribute by writing posts in their own languages, but since this multi-language environment can potentially affect the outcomes of content retrieval and analysis of those posts, the ultimate goal is properly separating the posts to obtain monolingual content. Therefore, language detection is an important part in facilitating content analysis of social media websites. In this paper, three approaches to language detection for short text messages have been developed and tested on Twitter data. Those approaches include: support vector machines (SVMs) and logistic regression, a probabilistic model based on modified Kneser-Ney smoothing, and a dictionary based approach. One important contribution of this paper is the extension of the probabilistic model to include additional information extracted from the Tweet objects. 
The first hypothesis that is examined here is that users mostly tweet in only a few languages, so storing the information about the languages connected to a particular user should result in improving the classification accuracy of the original model. Furthermore, the language that users choose as their user interface (UI) language should carry some information about the languages that those users tweet in as well, so it should be considered as relevant too. It is important to mention that this kind of meta-information extracted from Tweet objects is not just Twitter-specific, but it can be applied to texts from all the websites where there is certain access to user profiles, e.g. all social media websites, forums, blogs, etc, which makes this contribution widely applicable.

Many different approaches for tackling the language detection problem have been developed so far. Some of the best known models include the one of Cavnar and Trenkle \cite{Cavnar94n-gram-basedtext} popularized in the \texttt{textcat} tool, the \texttt{Chromium Compact Language Detector 2 (CLD2)} \cite{cld2}, originally extracted from the source code for Google Chromium's library by Michael McCandless and developed further by Dick Sites, and \texttt{langid} \cite{Lui:2012:LOL:2390470.2390475}, an off-the-shelf language identification tool by Lui and Baldwin. The Cavnar and Trenkle method uses a per-language character frequency model and classifies documents via their relative ``out-of-place'' distance from each language (see \cite{Cavnar94n-gram-basedtext} for more details). Variants on this method include Bayesian models for character sequence prediction \cite{Dunning94statisticalidentification}, dot products of word frequency vectors \cite{damashek95ngram}, and information-theoretic measures of document similarity \cite{Aslam:2003:IMD:860435.860545, Martins:2005:LIW:1066677.1066852}. \texttt{CLD2} and \texttt{langid} are both Naive Bayes classifiers, where \texttt{CLD2} probabilistically detects over 80 languages in Unicode UTF-8 text and for the mixed-language input returns the top three languages found for a given input and their approximate percentages of the total text bytes, while \texttt{langid} is trained on 97 languages over a naive Bayes classifier with a multinomial event model over a mixture of byte n-grams $(1 \le n \le 4)$ designed to be used off-the-shelf \cite{Lui:2012:LOL:2390470.2390475}. In the Results Section, the performance of \texttt{CLD2} and \texttt{langid} is compared to the performance of the methods developed in this paper. Additionally, kernel methods such as support vector machines (SVMs) were recently successfully applied to the same task \cite{ijcnn01tj, Lodhi:2002:TCU:944790.944799, Kruengkrai05languageidentification}, which motivated testing their performance on the short texts dataset from this paper. Recently, approaches based on deep neural network architectures are becoming increasingly common, with very promising results in language detection on speech data \cite{42538, 10.1371/journal.pone.0146917}. Even though these architectures are out of scope of the work done in this paper, they could be investigated in future work. 

However, the main difference between most of the approaches mentioned in the previous paragraph and the problem about to be tackled in this paper is that they are trained on large corpora with long, structured, well-written texts - e.g. the design target of \texttt{CLD2} are web pages with at least 200 characters (approximately two sentences) and it is not designed to do well on very short texts. The only algorithm from the ones mentioned for which its authors claim that it does well on short texts from the microblog domain is \texttt{langid}, which is the reason why we chose to compare its performance with ours in the Results Section of this paper. An interesting research by Baldwin and Lui \cite{Baldwin:2010:LIL:1857999.1858026} explores the impact of document length on language detection, with the conclusion that the performance accuracy improves significantly with increasing the document length. Therefore, in order for a method to achieve high accuracy results on short texts, it has to learn the particular characteristics of those texts, since this rather specific type of language is fairly difficult to match for methods trained on external corpora. The main advantage of the methods implemented in this paper compared to the methods targeted at long documents is the fact that they have been specifically developed to be able to recognize the language of short, noisy texts.
The problem of short texts language detection has been investigated by Nakatani Shuyo, who developed Language Detection with Infinity Gram (ldig). He reports quite impressive results of ``over 99\% accuracy for 19 languages'', on the corpus containing 700,000 labeled tweets. However, the drawback of ldig is that it is limited to Latin alphabet languages only, while some of the most common Twitter languages include Japanese, Korean, Chinese, etc. Furthermore, his analysis is limited to texts longer than 3 words, which is often not the case when dealing with Twitter data. Another example of short text language detection is done in \cite{Carter:2013:MLI:2447287.2447296} and it relies on the results of the before mentioned \texttt{textcat} tool. Similarly to Shuyo's work, the results presented here are on a dataset containing only 5 Latin alphabet languages, which is considered not nearly enough to declare having a reliable short texts language detection algorithm task solved.  In this paper, we try to overcome the difficulties accompanying short texts language detection compared to the longer texts on the one hand, while at the same time including the non-Latin alphabet languages in the dataset, since they are considered to be an important subset of languages which should not be excluded.

\subsection{Datasets}
\begin{figure}[!ht]
    \centering
        \includegraphics[width=0.8\textwidth]{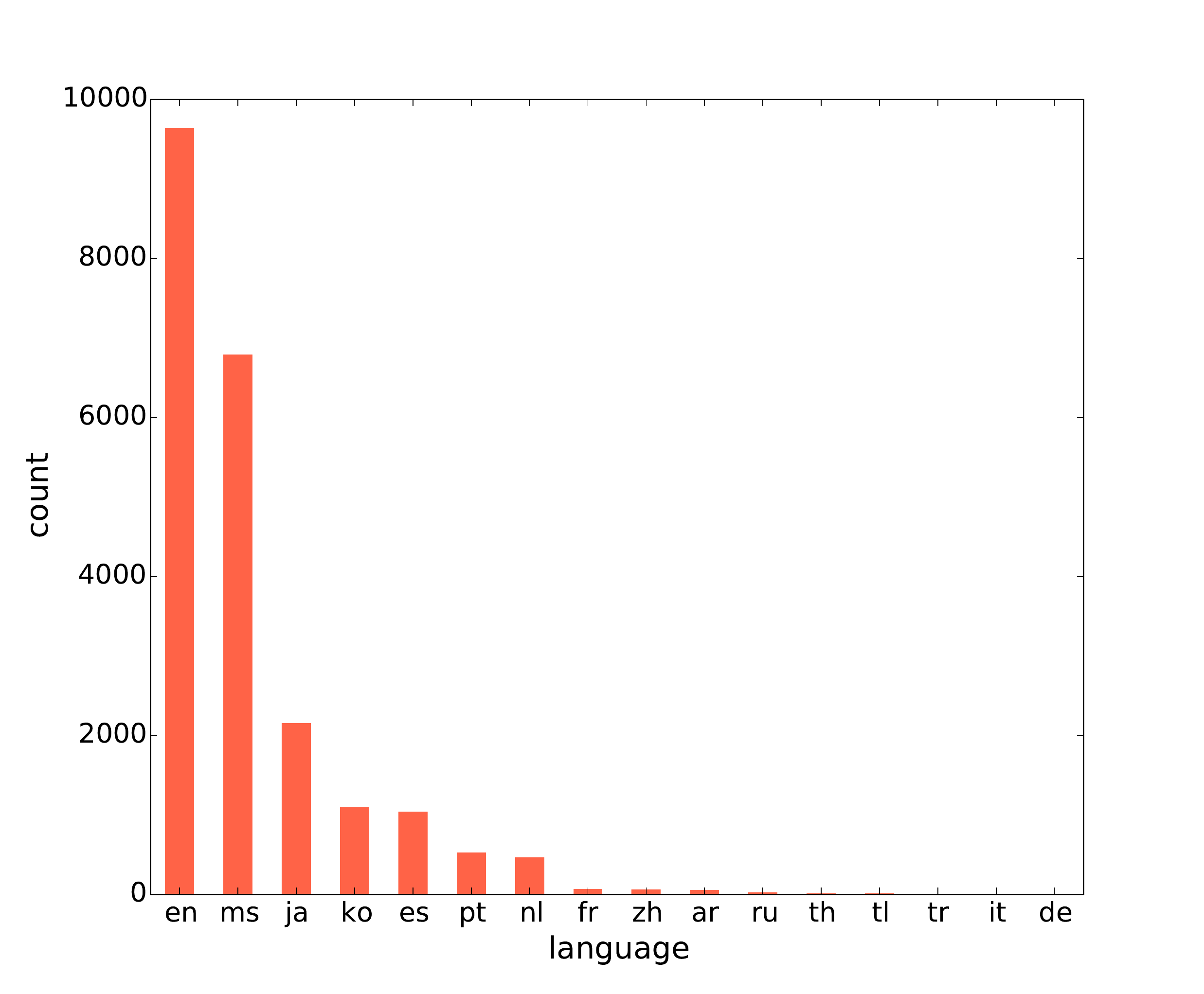}
    \caption{Distribution of languages in the dataset}
    \label{fig:langdist}
\end{figure}

The dataset that is used for this task consisted of .json files in which the \texttt{Tweet} objects are stored. Each \texttt{Tweet} object contains different types of relevant information about its nature, such as the unique identifier of the tweet itself, the text that it contains, the information about its author, time and location at the point of creation, etc. However, only parts of this information are considered relevant for the classification task.
The files that are used contain tweets collected using the Twitter API in April 2012, where in total around 22,000 tweets in 16 different languages are randomly collected at different time points during two days. Languages appearing in less than 3 tweets are discarded and due to insufficient domain knowledge, Indonesian and Malay are grouped together to one language. The language distribution across the dataset is shown in Fig. \ref{fig:langdist}. When collecting the data, we complied with the Twitter's Terms of Use.
As expected, almost half of the total number of tweets are written in English. The languages following English by the number of tweets are Malay, Japanese, Korean, Spanish, Portuguese, and Dutch.
Even though the distribution of the number of tweets per language in the dataset is highly skewed, it corresponds quite well to the distribution of Twitter languages given in \cite{journals/corr/abs-1212-5238}.

To make the language labeling of the tweets easier and reduce the manual work, the open-source language detection library \texttt{Chromium Compact Language Detector 2 (CLD2)} is used. The results obtained by \texttt{CLD2} are then manually checked and all the wrongly assigned labels are corrected in order to obtain a clean dataset and avoid repeating the same mistakes that \texttt{CLD2} made. Finally, the column indicating the tweet language is added to the existing .csv file. However, since only around 8,000 tweets are obtained using this rather time-consuming approach, additional tweets are collected from the users with the user ID already existing in the dataset, assuming the majority of users would tweet only in one or two languages. Therefore, the language from the tweet already present in the dataset is assigned to all the remaining tweets from the same user. To prevent possible mistakes when using this approach, all the newly collected data is later again manually rechecked.

\section{Methods}

The general task of language detection is to predict, for a given text $t$, the language $l$ in which the text is written.
A typical naive approach to solving the language detection task would be to show the text to a certain language expert, who would then decide on a language the text is written in. However, that would require many different language experts for each of the languages. This solution becomes even more problematic if the database of texts is not static but it is changing over time, where scalability becomes an issue. Therefore, a machine learning approach is needed. In the machine learning approach to solving the language detection problem, we are given a certain amount of \emph{data} (a set of texts in different languages) and the \emph{labels} (languages to which those texts belong). The labels have previously been assigned to the data by some form of annotation procedure. Even though the need for human language experts still exists in the annotation step, once the initial amount of data is labeled, the algorithm does the rest of the work. Having the labels and not just the raw texts makes this a \emph{supervised learning problem}, in which each example is a pair of the input object and the target value. The goal of supervised learning is to predict the correct output value for each input object.

\subsection{Preprocessing} \label{preprocessing}
 
The preprocessing task is usually the first part in a machine learning document processing pipeline, preceding the extraction of features from the data. In this paper, preprocessing included editing the tweet texts and assigning them the corresponding language labels. The text editing in general consists of cleaning the texts, removing all the information considered to be unnecessary for the task, and transforming all the texts into the same, mutually comparable form. As the first preprocessing step, all the links and expressions of addressing a particular user (the \emph{@user\_name} form) are removed from every tweet text using simple regular expressions, as they are considered irrelevant for the differentiation between languages. In addition, all the emoticons are removed too, since they maintain the same form across the languages. The text is then converted to lowercase, all multiple white spaces are trimmed, and all the punctuation marks are removed. This procedure transformed all the texts into an equal format, to improve the accuracy when performing their mutual comparisons.

\subsection{Feature Extraction}\label{features}

In this paper, two different types of features are extracted from the tweet texts, depending on the classifier used: character n-grams and bag-of-words features.
\emph{Character n-grams} can be described as all character substrings of length $n$ in the given text. On the other hand, the \emph{bag-of-words} features are defined as an unordered collection of words in the text. Whenever possible, the character n-gram feature model is chosen over the bag-of-words model, which is justified by the specific type of language used in the dataset. Namely, character n-grams model is more resilient against misspellings, abbreviations, acronyms, and word derivations than the bag-of-words, since it does not strictly impose the splitting of texts by white spaces.

For SVM and logistic regression classification, character n-grams are chosen as the appropriate feature type. After extracting the n-grams, the next step is to transform this collection of features into numerical feature vectors, which is a standard step before applying most of the machine learning algorithms to text data. This task can be done in many ways - from the simplest one of having a binary indicator whether a particular n-gram appeared in a text to counting the occurrences of that n-gram in a text and optionally applying different kinds of normalizations to those counts. 

In order to choose the most appropriate feature type for this task, the evaluation of different $n$ values and normalization types is done on a held-out development set by performing a 5-fold cross-validation procedure. For every combination of $n$ values (the values of $n =\{2, 3, 4\}$ are considered reasonable for short Twitter texts) and normalization types (\emph{tf-idf} and length normalization are examined here), the micro- and macro-averaged F1-scores are computed using the \texttt{scikit-learn} software \cite{scikit-learn}.
For both classifiers, the value of $n=2$ with the \emph{tf-idf} weighting is chosen as the best feature type, since it slightly outperformed all the other parameter combinations.

In the probabilistic approach, character n-grams are again chosen as the best suited feature type. Due to the use of the modified Kneser-Ney smoothing \cite{Chen:1996:ESS:981863.981904} and its recursive nature, character 1-4-grams are chosen as the appropriate features. Namely, the Kneser-Ney probabilities of the higher order n-grams are computed using the probabilities of the lower order n-grams. Limiting the order of the character n-grams to $4$ seemed as the most reasonable choice here, due to tweets being too short to extract features longer than 4-grams. No normalization is performed in this approach, since the Kneser-Ney algorithm is designed to work directly on n-gram counts.

In the dictionary based approach, the \emph{bag-of-words} feature model is the only possible feature model, since this approach relies on matching the words from a dictionary with the words in the text, so the features necessarily need to be whole words. Therefore, the character n-gram model is not considered here.

\subsection{Classification} \label{classification}

\paragraph{SVM} \label{svm}

\emph{Support vector machines (SVMs)} are supervised learning models, used mostly for classification and regression problems. In the next paragraphs, the SVM classification is described for the case of only two classes for simplicity, since multi-class classification is just an extension of that model \cite{Crammer:2002:AIM:944790.944813}. The multi-class support in this paper is handled according to the one-vs-one scheme. SVM classification is focused on trying to maximize the \emph{margin}, i.e. the distance of the data points of both classes from the decision boundary based on structural risk minimization \cite{Vapnik:1995:NSL:211359}. One way to achieve this is by solving the dual optimization problem \cite{schoelkopf01kernelbased}. The dual optimization problem is defined as follows:

\begin{equation*}
 \begin{aligned}
 & \underset{\alpha}{\text{max}}
 & & \sum_{i=1}^n \alpha_i - \frac{1}{2}\sum_{i, j} \alpha_i \alpha_j y_i y_j k(\mathbf{x}_i, \mathbf{x}_j) \\
 & \text{subject to} 
 & & 0 \leq \alpha_i \leq C, \; i = 1, \ldots, n \\
 & 
 & &  \sum_{i=1}^n \alpha_i y_i = 0
 \end{aligned}
\end{equation*}
where $\mathbf{x}_i$ are the feature vectors and $y_i$ are the corresponding class labels. The function $k(\mathbf{x}_i, \mathbf{x}_j)$ is the so-called \emph{kernel} function, which describes the similarity between two documents and allows the extension of SVMs to nonlinear problems. The parameter $C$ is a regularization constant, which allows for some points in the training set to be misclassified, in order to avoid overfitting. All data points with $\alpha_i>0$ are the so-called \emph{support vectors}, i.e. those data points that lay on or inside the margin. The $\alpha_i$-s are typically equal to 0 for most of the documents considered, which makes SVMs extremely efficient: when assigning the label to a new data point, only those documents have to be considered which have support vectors larger than 0. After the training phase is completed, a new data point is classified according to the following expression \cite{schoelkopf01kernelbased}:
\begin{equation*}
f(\mathbf{x}) = \text{sign}\left(\sum_{i=1}^n y_i \alpha_i k(\mathbf{x}, \mathbf{x}_i) + b\right)
\end{equation*}
where $b$ is the bias term. One of the most important points to consider when choosing an SVM as a classification method is the choice of the corresponding kernel function, since the appropriate choice of the kernel function significantly influences the classification accuracy. In this paper, the \emph{linear kernel} is chosen, since it performs well in the cases where the dimensionality is much higher than the number of data points. Additionally, computing the linear kernel requires less computational cost than computing any of the other kernel functions (e.g. rbf, polynomial, etc.), since a linear kernel is just a simple dot product in the feature space. Therefore, training an SVM with a linear kernel is faster than with any other kernel, particularly when using a dedicated library such as \texttt{LibLinear} \cite{HsuLibsvmTutorial2003}. Finally, most of the text classification problems are linearly separable \cite{Joachims:1998:TCS:645326.649721}, so no other kernels except for the linear are needed. SVM is chosen as one of the models in this work because of its many advantages \cite{scikit-learn}: it is effective in high dimensional spaces, it is still effective even in cases where the number of dimensions exceeds the number of samples (usually the case with categorization of text documents), it uses only a subset of training points in the decision function, so it is also memory efficient, and it is unlikely to overfit, since the ratio of number of data points and effective dimensions is typically high \cite{journals/jmlr/BraunBM08}, given that an appropriate regularization term is used.

\paragraph{Logistic Regression} \label{logreg}

\emph{Logistic regression}, despite having the word ``regression'' as part of its name, is a linear model for \emph{classification} rather than regression \cite{Bishop:2006:PRM:1162264}. The logistic regression classification paradigm is described here for the two class case only. It is a type of probabilistic statistical model, where the probabilities describing the possible assignments to different classes are modeled using the logistic function \cite{scikit-learn}, which is defined as:

\begin{equation*}
 \begin{aligned}
P[y_i=+1|x_i,w] = \frac{e^{w^\top x_i}}{1+e^{w^\top x_i}}
 \end{aligned}
\end{equation*}
where $y\in\{-1,+1\}$ is the assigned class label, $x_i$ is the data point, $w$ is the regression coefficient, and $P[y=+1|x_i,w]$ is the probability of $x_i$ being drawn from the positive class. A new data point $x_i$ gets assigned to a class with the highest probability. As an optimization problem, two-class L2-penalized logistic regression minimizes the following cost function:

\begin{equation*}
 \begin{aligned}
& \underset{w}{\text{min}}
& & \frac{1}{2}w^\top w + C \sum_{i=1}^n{log(e^{-y_i{X_i}^\top w}+1})
 \end{aligned}
\end{equation*}
 where $\frac{1}{2}w^\top w $ is the L2-regularization and $C$ is the inverse regularization constant. The reasons for using logistic regression in this work are: its simplicity - it creates a linear decision boundary, it is effective in high dimensional spaces, and it is unlikely to overfit when appropriate regularization term is chosen.

 \paragraph{}
 After extracting the character n-grams features as described in the Feature Extraction Section, the obtained feature matrix containing the \emph{tf-idf} features and the label vector containing the corresponding class labels are split into training and test parts, which is repeated in a 5-fold \emph{cross-validation} procedure. After being trained on the training data using the \texttt{scikit-learn} implementation of the two classifiers, the learned models are then applied to the test data. 
During the classification procedure, special treatment is given to texts containing the characters belonging to Thai, Arabic, Korean, Japanese, and Chinese language, due to the very large number of different characters present in each of those languages and as a result, classifiers performing poorly on texts belonging to those languages. Therefore, SVM and logistic regression classifiers are not trained on texts coming from those languages, but a specific method is applied to determine from which language the text originated from. For this reason, a designated \emph{threshold} value is determined experimentally on the held-out development set in order to check if a test text belongs to one of the languages mentioned. If the special characters make more than the specified percentage of the text length, the text gets assigned the language label to which those special characters belong. The biggest challenge here is the differentiation between Japanese and Chinese, since they both use Kanji characters. However, the number of Kanji characters used in Chinese is much larger than in Japanese and Hiragana and Katakana are specific to Japanese only, which enables successful separation of the two languages.

\paragraph{Dictionary Based Approach}\label{dictbased}

The dictionary based method is by far the simplest one of all the methods tested - it is based on having the dictionary of all possible words for each language, comparing the words in the text with the words in each of the dictionaries and counting the number of hits per text and per language. The winning language label  for each text is the one with the highest number of hits.
The primary advantage of the dictionary based approach is its simplicity - it does not require a training phase and the algorithm itself is very easy to implement. However, usually it cannot compete with other more powerful methods, as shown in the Results Section. Additionally, it uses a lot of memory by saving all the dictionaries, which is efficiently dealt with by using \emph{bloom filters} for each dictionary rather than iterating over every dictionary document.

The first step of the dictionary based approach is to download the dictionary of all the words for each of the languages. For that, the \emph{GNU Aspell} dictionaries are used. The dictionary files are then preprocessed in order to include only one word per line and stored as bloom filters, due to time and space efficiency reasons. A bloom filter is a space-efficient probabilistic data structure used to test whether an element is a member of a set, where false positive matches are possible while false negatives are not. In other words, it is
possible to conclude if an element is “possibly in set” or “definitely not in set”.
Each of the words in a single tweet text is checked against every bloom filter and the number of alleged hits is counted accordingly. The text is then assigned to the language with the highest number of hits. Some Asian languages (Thai, Korean, Japanese, and Chinese) are handled the same way as described in the previous paragraph, since there are no GNU Aspell dictionaries for those languages.

\paragraph{Probabilistic Model Based on Modified Kneser-Ney Smoothing}\label{probmodel}

The probabilistic model algorithm implemented in this work outputs a vector of probabilities for a certain text belonging to each of the languages present in the data set. The language with the highest probability assigned gets chosen as the class label. The algorithm is based on \emph{modified Kneser-Ney smoothing}, which is a slightly altered version of Kneser-Ney smoothing, proven to outperform the original version \cite{Chen:1996:ESS:981863.981904}. Kneser-Ney smoothing makes use of absolute discounting by subtracting a fixed value from the lower order terms to omit n-grams with lower frequencies, i.e. it takes into account the frequency of unigrams in relation to possible higher order n-grams in which those unigrams are contained. Additionally, this smoothing results in assigning a probability value greater than zero to all n-grams which are not appearing in the training set but are present in the test set. Modified Kneser-Ney smoothing computes the conditional probability $p_{KN}(w_i|w_{i-n+1}^{i-1})$ for each n-gram in every text, where $w_i$ is the $i$-th n-gram in the text and $w_{i-n+1}^{i-1}$ are the preceding $n-1$ n-grams. The probability value for the whole text $p(t)$ is then computed as:

\begin{equation*}
 \begin{aligned}
p(t) = \prod_{i=1}^{l+1}{p_{KN}(w_i|w_{i-n+1}^{i-1})}
\end{aligned}
\end{equation*}
where $l$ is the number of n-grams in the text. The conditional probability $p_{KN}$ is then defined recursively as:

\begin{equation*}
 \begin{aligned}
p_{KN}(w_i|w_{i-n+1}^{i-1}) &= \frac{c(w_{i-n+1}^{i}) - D(c(w_{i-n+1}^{i}))}{\sum_{w_i}{c(w_{i-n+1}^{i})}} + \gamma(w_{i-n+1}^{i-1})p_{KN}(w_{i-n+2}^{i-1})
\end{aligned}
\end{equation*}
where $c$ denotes the count, $\gamma(w_{i-n+1}^{i-1})$ is the scaling factor to make the distribution sum to 1 and $D$ is the discount factor where:

\begin{equation*}
 \begin{aligned}
 D(c) =
  \begin{cases}
  0 & \text{if } c=0 \\
  D_1 & \text{if } c=1 \\
  D_2 & \text{if } c=2 \\
  D_{3+} & \text{if } c \geq 3
  \end{cases}
\end{aligned}
\end{equation*}
i.e. instead of using a single discount $D$ for all non-zero counts as in Kneser-Ney smoothing, three different parameters $D_1$, $D_2$, and $D_{3+}$ are applied to n-grams with one, two, and three or more counts, respectively. To make the distribution sum to 1, $\gamma(w_{i-n+1}^{i-1})$ is defined as:

\begin{equation*}
 \begin{aligned}
\gamma(w_{i-n+1}^{i-1}) = \frac{D_1N_1(w_{i-n+1}^{i-1}\cdot) + D_2N_2(w_{i-n+1}^{i-1}\cdot) + D_{3+}N_{3+}(w_{i-n+1}^{i-1}\cdot)}{\sum_{w_i}{c(w_{i-n+1}^{i})}}
\end{aligned}
\end{equation*}
where $N_{x+}$ stands for the number of words that have $x$ or more counts and $\cdot$ is a free variable, on which has been summed over. 
The estimates for the optimal discount values $D_1$, $D_2$, and $D_3$ are computed as a function of training data counts \cite{finke}:
\begin{equation*}
 \begin{aligned}
 & D_1 = 1 - 2Y\frac{n_2}{n_1} \\
 & D_2 = 2 - 3Y\frac{n_3}{n_2} \\
 & D_{3+} = 3 - 4Y\frac{n_4}{n_3} 
\end{aligned}
\end{equation*}
where $Y=\frac{n_1}{n_1+2n_2}$ and $n_x$ is the number of n-grams $n$ appearing $x$ times in the training data. It is important to mention that if the smoothing term was omitted, the probability value of the whole text $p(t)$ would be 0 for each text that contains an n-gram present in the test data but which never appeared in the training data.
After computing the probabilities for each text and each language, the text gets assigned the language label with the highest probability score. The advantage of this probabilistic model over the other methods is that it takes into account the n-grams with zero-counts by smoothing the probability function, which should in turn lead to higher accuracy. However, the method also has certain drawbacks, such as its complexity and high space and time usage due to its many recursive calls.

In this approach, the training phase consists of counting the number of occurrences of a specific 1-4-gram in texts of each language. This dictionary of n-gram counts is then fed into the test phase. The test phase consists of iterating over every n-gram of each text from the test data and computing the modified Kneser-Ney probability of that n-gram belonging to a certain language. The probability of the whole text belonging to a certain language is calculated by multiplying the probabilities for all the n-grams contained in that text. Finally, the language with the highest probability is chosen. One important fact here is that special handling for non-Latin languages that is used for SVM, logistic regression, and the dictionary based approach is not used here, since the probabilistic model performed well on texts belonging to those languages.

\subsection{Including Additional Information}\label{addinfo}

The most important contribution and the biggest novelty of this paper compared to previous work done in the language detection field is the extension of the probabilistic model to include additional personalized user information. Due to the fact that the output of the probabilistic model is in the form of a probability distribution over different languages, additional information can be added to the model in order to improve the predictions. Two types of information are investigated here and added to the original model and their impact on the predictions is evaluated. This information includes prior information about the language usage by a particular user and information about the user interface language.

First, a \emph{prior} frequency distribution for choosing a specific language is defined for each user. This distribution is chosen to be uniform at the beginning, with an experimentally determined value, as described later in the Results Section. For every new text in the test data, the chosen language is no longer determined by looking only at the probability distribution given by the classifier as before, but at the product of that distribution with the prior user-specific distribution. This user-specific prior distribution gets adjusted each time after observing a new text and determining its language by increasing the count for that language. This way, the \emph{user-specific evidence accumulation} is included in the model. Since it is assumed that an average user tweets in only a few languages, this adjustment of the model is believed to help improve the overall results. The adjustment should be especially important in the cases of texts with high uncertainty in the results gained directly from the original classifier. This is best illustrated by an example - if a user frequently tweeted in language A but the Kneser-Ney probability for the current tweet is not large enough to decide for language A and it is really close to the probability of a language B, the user data accumulation will help decide for the more frequently used language, in this case A. On the other hand, if the Kneser-Ney probability for language B is much higher than the one for language A, the algorithm is still going to decide for the language B, if an appropriate prior distribution is chosen. The value to which the flat prior distribution is initialized decides in this case how important a new data point is - if a prior is set to a low value, each new data point influences the distribution greatly; if it is on the contrary set to a high value, a lot of new data has to be seen to significantly change the distribution. 

Furthermore, it is assumed that the user interface (UI) language chosen by the user should carry additional information about the language the user tweets in, i.e. those two languages should be equivalent in some cases. Therefore, it has been decided to include the \emph{information about the UI language} in the original model as following: the prior distribution is no longer set to a uniform value at the beginning of the classification procedure, but the value for the UI language is increased, again by an experimentally determined amount.

\section{Results}
In this first part of this Section, we give a brief introduction to accuracy measures used to compare performances of the classifiers. To gain better insight into the performance of different methods, their prediction results are compared in the second part of this Section. In addition, the effect of adding additional information to the model other than classifying based just on character n-grams is evaluated in the next Subsection. The best performing method is then chosen and those results are compared to the \texttt{CLD2} and \texttt{langid} results, which is described in the last Subsection. 

\subsection{Accuracy Measures}

A good classifier is defined as one for which the number of true positives (TP) and true negatives (TN) is high, while at the same time keeping the number of false positives (FP) and false negatives (FN) low. To sum up those outcomes, two different accuracy measures were used in assessing the classifiers performance: micro- and macro-averaged F1-score. In order to understand the F1-score, precision and recall need to be explained first.
\emph{Precision} is defined as:
\begin{equation*}
 \begin{aligned}
\text{precision}=\frac{\mbox{TP}}{\mbox{TP + FP}}
 \end{aligned}
\end{equation*}
and it measures the ability of the classifier not to assign a sample to the class to which it does not belong.
\emph{Recall} is defined as: 
\begin{equation*}
 \begin{aligned}
\text{recall}=\frac{\mbox{TP}}{\mbox{TP + FN}}
 \end{aligned}
\end{equation*}
and it measures the ability of the classifier to find all the samples that belong to that class (both assigned to it and the ones not assigned).
\emph{F1-score} is defined as: 
\begin{equation*}
 \begin{aligned}
\text{F1-score}=\frac{\mbox{2 $\cdot$ precision $\cdot$ recall}}{\mbox{precision + recall}}
 \end{aligned}
\end{equation*}
and it is the weighted average (harmonic mean) of precision and recall. The \emph{micro-averaged} F1-score calculates the metrics globally by counting the total number of TPs, FPs, and FNs, while the \emph{macro-averaged} F1-score calculates the metrics for each label and finds their unweighted mean, not taking label imbalance into account \cite{scikit-learn}. Because of the large difference in sample sizes between the languages in the dataset used in this paper, the difference between the micro- and macro-averaged F1-scores is expected to be large as well. The lack of training data for some languages (e.g. Turkish, Italian, German, see Fig. \ref{fig:langdist})does not allow the classifier to learn the correct representation for those languages. Therefore, the F1-scores for those labels are expected to be low, which will then affect the macro-averaged F1-score in a negative way.

\subsection{Performance Comparison of Different Methods}\label{methodcomparison}

In order to assess the quality of the SVM and logistic regression predictions, a 5-fold cross-validation procedure is performed on both of the classification methods and the presented results are in the form of mean micro- and macro-averaged F1-scores and their standard deviations over the folds. Additionally, SVM and logistic regression are combined together to form an \emph{ensemble learning} method, since it is assumed that this will increase the classification accuracy, as suggested in \cite{Opitz99popularensemble} and \cite{Rokach:2010:EC:1713727.1713730}. 
In the ensemble learning, the confidence scores for choosing a specific language are in fact the probabilities gained from both of the classifiers. Precisely, if the predicted language for a certain text differs across the classifiers, the label is taken from the classifier which yields a higher confidence score. 
However, since the \texttt{scikit-learn} implementation of the SVM classifier does not implicitly include probability scores, those are obtained with the use of Platt scaling \cite{Platt99probabilisticoutputs} by setting the \emph{probability} parameter to True. 
The corresponding micro- and macro-averaged F1-scores can be seen in Table \ref{table:classifiers}. Contrary to the previous hypothesis, the SVM itself outperforms the ensemble classifier both in micro- and macro-averaged F1-scores. Obviously, the logistic regression fails to capture some information about the correct decision boundary between the language classes, while still having high confidence about the predictions. In assessing the performance accuracy of the probabilistic model based on modified Kneser-Ney smoothing, the 5-fold cross-validation procedure is again performed. It is shown in Table \ref{table:classifiers} that the probabilistic model with Kneser-Ney smoothing significantly outperforms the traditional classifiers, presumably due to its smoothing of the n-gram counts distribution to account for the zero-probabilities n-grams. It is important to mention that no smoothing is done with the traditional classifiers, since that would make the feature matrices no longer sparse, which would in turn result in huge computational costs. 

\begin{table}[!ht]
\small
\begin{center}
\begin{tabu}{ l | c | c }
  method & micro-averaged F1& macro-averaged F1\\ 
  \tabucline[2pt]{-}
  SVM & $96.92 \pm 0.36$ & $73.11 \pm 2.25$ \\
  \hline
  logistic regression & $96.72 \pm 0.33$ & $72.14 \pm 3.51$ \\
  \hline
  ensemble learning & $96.86 \pm 0.34$ & $72.83 \pm 3.84$ \\
   \hline
  probabilistic model & $\mathbf{98.25 \pm 0.12}$ & $\mathbf{74.12 \pm 3.07}$\\
   \hline
  dictionary based & $89.14 \pm 0.01$ & $41.78 \pm 0.01$\\
\end{tabu}
\vspace{2mm}
\caption {Micro- and macro-averaged F1-scores for different methods}
\vspace{-4mm}
\label{table:classifiers}
\end{center}
\end{table}
\noindent To assess the dictionary based method, no training phase is necessary, since the algorithm just compares the words from the tweets with the words in the provided dictionaries. The performance of the algorithm is still measured on the same test sets as the other methods, in order to ensure fair comparison. It is obvious that the dictionary method performs a lot worse than the other two methods, which confirms the initially expected outcome. Some possible reasons for that could be that the texts in the dataset are very short (a tweet is limited to 140 characters), which is not enough to gain a different number of hits for different languages. They are also filled with spelling mistakes, which changes the original words so they don't match exactly the ones in the dictionary. Additionally, this outcome also shows the weakness of the bag-of-words feature model compared to the n-gram model used in other approaches.

\subsection{What if we add additional information?}\label{additionalinfo}

\begin{figure}[!ht]
    \centering
        \includegraphics[width=0.9\textwidth]{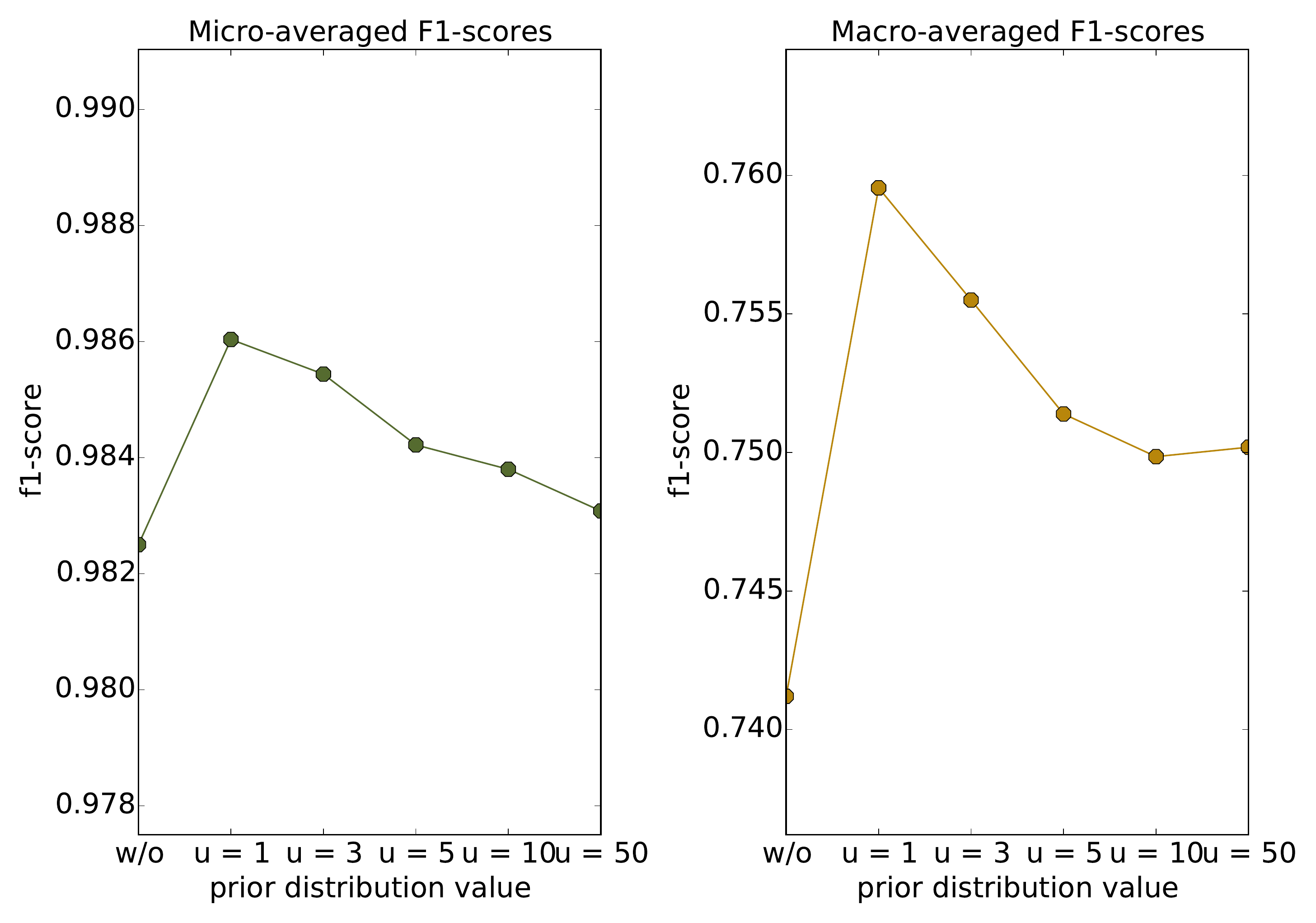}
    \caption{Micro- and macro-averaged F1-scores for different prior distribution values}
    \label{fig:priors}
\end{figure}

\begin{figure}[!ht]
    \centering
        \includegraphics[width=0.9\textwidth]{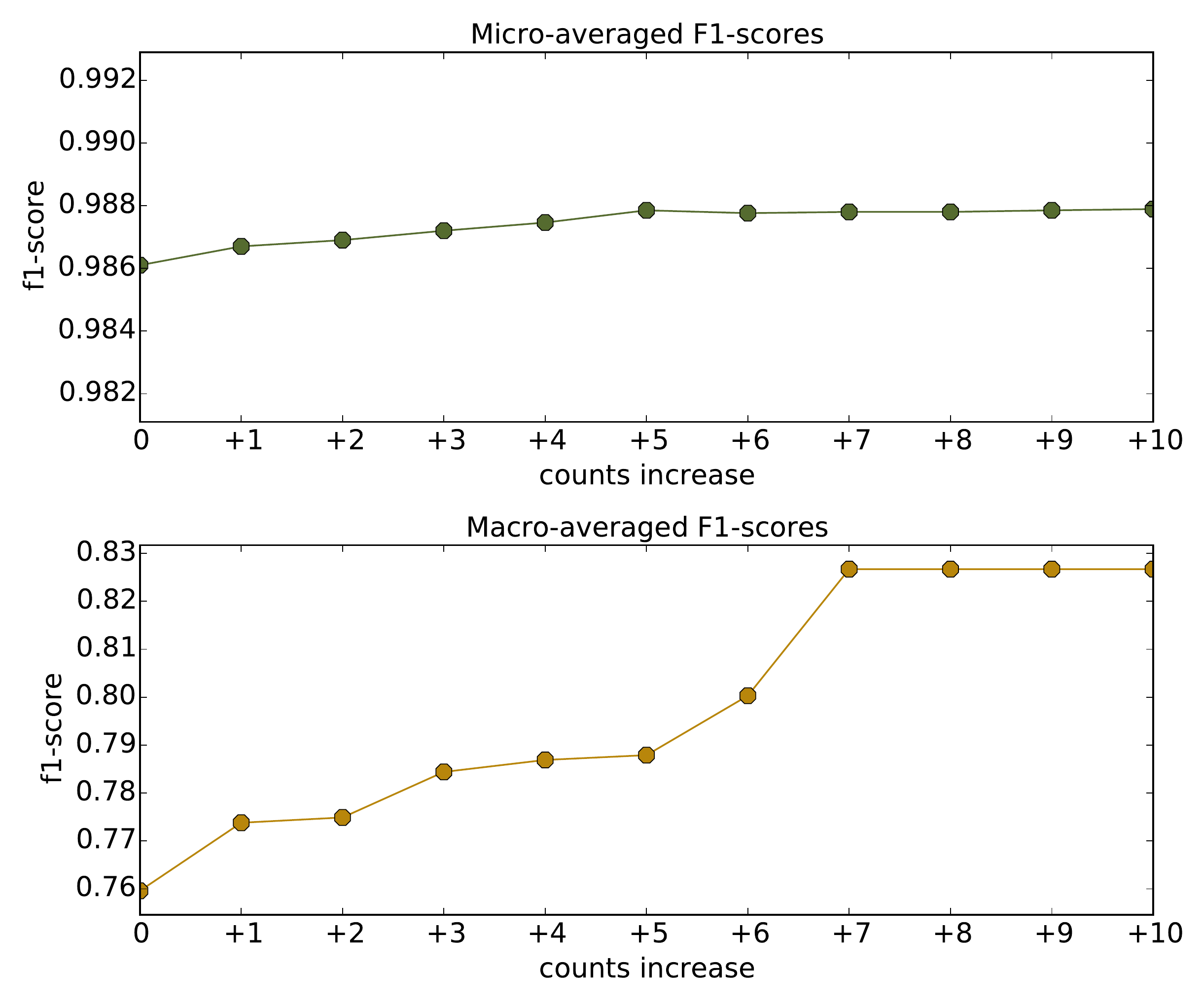}
    \caption{Micro- and macro-averaged F1-scores for assigning different importance values to the UI language information}
    \label{fig:uilang}
\end{figure}

Under the hypothesis that the classification accuracy can be improved by adding additional information to our model other than only n-gram frequency counts, different approaches are tested together with the probabilistic model, since that is the model that yielded the best results in comparison with other methods. First, the effect of accumulating the data per user is implemented. For each new tweet, we extract its user ID and if this user ID already appeared in the dataset before, we increment the count for the number of tweets in that language for that user. However, if this is the first tweet by that user, the prior distribution of tweet counts per language is set to a uniform distribution. The exact value of the uniform distribution is determined experimentally on a held-out development set and different values $\{1, 3, 5, 10, 50\}$ are tested to analyze the influence that the height of the prior has on the F1-scores. Fig. \ref{fig:priors} shows the micro- and macro-averaged F1-scores for different prior distribution values. The lower the prior distribution value is, the better the micro- and macro-averaged scores get. Therefore, the best performing model is the one where the prior distribution is flat with the value $1$ and it achieves the micro-averaged F1-score of $98.61 \pm 0.08$ and the macro-averaged score of $75.96 \pm 2.72$. Additionally, it is obvious that all the results including the user-specific information perform better than the one without it, which confirms the hypothesis that including user-specific information is a valuable extension of the model.
The second hypothesis includes the users' UI language. The UI language is included in the uniform prior distribution as following - the count for that language is increased compared to the counts for other languages. The exact increase amount is again determined experimentally, where the values of $\{+1, +2, +3, +4, \dots, +10\}$ are tested on the held-out development set, while the counts for all the other languages are set to $1$, due to the results presented earlier in Fig. \ref{fig:priors}. All of the above mentioned values increased the classification accuracy compared to the procedure where the UI language information is not included in the model, as it can be seen in Fig. \ref{fig:uilang}. The best results are achieved when using the count increase value of $7$.
 Compared to the model without the added UI language information, a very significant increase in classification accuracy of $6.71$ is achieved when looking at the macro-averaged F1-scores. The potential reason for that may be that the information about the UI language influenced the most those texts for which not enough training data is available, which improved the overall macro-averaged score. Regarding the micro-averaged scores, an improvement of $0.18$ has been achieved compared to the model without the UI language information.

\begin{table}[!ht]
\small
\begin{center}
\begin{tabu}{ l | c | c }
  method & micro-averaged F1& macro-averaged F1\\ 
  \tabucline[1.5pt]{-}
  prob. model & $98.25 \pm 0.12$ & $74.12 \pm 3.07$\\
  \hline
  prob. model \& evidence acc. & $98.61 \pm 0.08$ & $75.96 \pm 2.72$ \\
  \hline
  prob. model \& evidence acc. \& UI & $\mathbf{98.79 \pm 0.15}$ & $\mathbf{82.67 \pm 3.54}$ \\
\end{tabu}
\vspace{2mm}
\caption {Micro- and macro-averaged F1-scores after adding user-specific information to the probabilistic model}
\label{table:summary}
\end{center}
\vspace{-4mm}
\end{table}

In conclusion, the chosen classification method is the probabilistic model with modified Kneser-Ney smoothing, where the feature model is based on character n-grams with addition of user-specific information in form of evidence accumulation and UI language information. With that method, the classification accuracy of $\mathbf{98.79 \pm 0.15}$ is obtained in the micro-averaged F1-scores and $\mathbf{82.67 \pm 3.54}$ in the macro-averaged F1-scores. The performance overview of all the approaches tested with the probabilistic model is shown in Table \ref{table:summary}.

\begin{figure}[!ht]
    \centering
        \includegraphics[width=0.8\columnwidth]{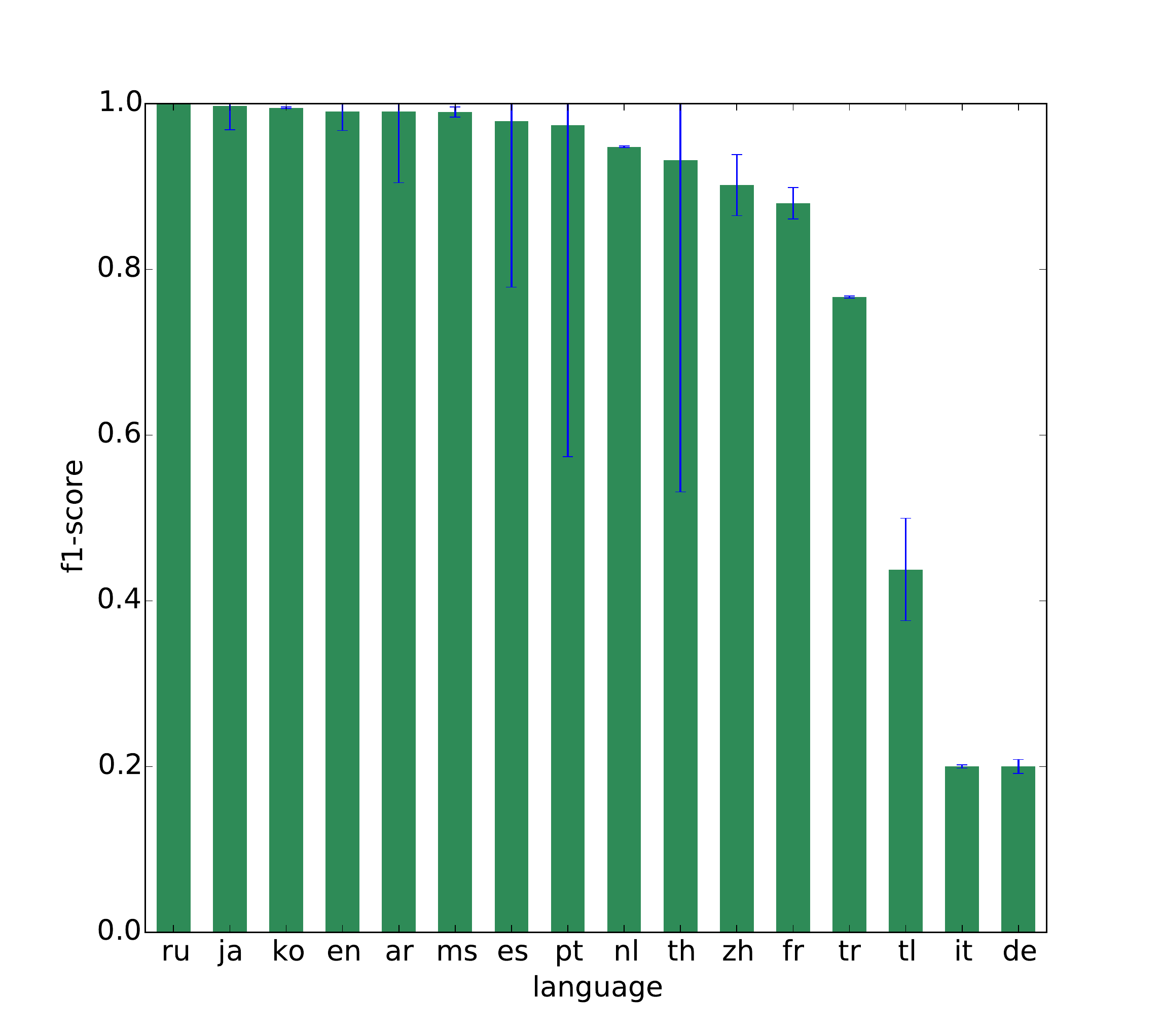}
    \caption{The distribution of F1-scores across different categories}
    \label{fig:f1perlang}
\end{figure}

Additionally, the distribution of F1-scores across different languages is plotted for the best performing method in Fig. \ref{fig:f1perlang}.
By comparing the results presented in Fig. \ref{fig:f1perlang} with the number of samples per category shown in Fig. \ref{fig:langdist}, it can be seen that the 4 languages that had the least samples (Turkish, Tagalog, Italian, German) are also the ones with the lowest F1-scores. This confirms the conjecture that not having enough training data affects the classification accuracy. As expected, languages with a large number of samples (English, Malay, Spanish, Portuguese, Dutch) achieve high F1-scores as well. However, it is interesting to notice that very good results are obtained for the texts belonging to the non-Latin languages such as Russian, Japanese, Korean, Arabic, and Thai. This is an indication that the representation of those categories is easily learned by the model, due to the special type of characters used in those languages. The only non-Latin language where the results are not so high is Chinese, possibly due to the large overlap between the characters used in Japanese and Chinese.

\subsection{Performance Comparison with \texttt{CLD2} and \texttt{langid}}\label{cld2_langid_comparison}
The goal of this Section is to compare the performance of the best method from the previous Section with the performance of the \texttt{Chromium Compact Language Detector 2} and the \texttt{langid} tools. The categorization is done on both raw tweets and the preprocessed ones, where the preprocessing procedure is the same as the one described in the Preprocessing Section of this paper. The language chosen as the predicted language by \texttt{CLD2} is the one with the highest confidence, as outputted by the \texttt{CLD2} algorithm, while \texttt{langid} outputs the language with the highest confidence only. The obtained results are visible in Table \ref{table:cld2_langid}. The results obtained by \texttt{CLD2} improve significantly when applying the preprocessing methods developed in this paper compared to the results on the raw data, but they are still considerably worse than the results achieved by our algorithm. The macro-averaged F1-scores differ slightly more between the classifiers than the micro-averaged ones. 
The reason for that may lie in the fact that the probabilistic model outperforms \texttt{CLD2} and \texttt{langid} for some languages for which there is not a lot of data available, an aspect that we attribute to the usage of prior information. 
Results achieved by \texttt{langid} do not depend much on preprocessing the data, but they are significantly worse than the results achieved by our algorithm in both micro- and macro-averaged F1-scores, even though the authors claim it should perform well across different domains.

\begin{table}[!ht]
\small
\begin{center}
\begin{tabu}{ l | c | c }
  method & micro-averaged F1& macro-averaged F1\\ 
 \tabucline[1.5pt]{-}
  cld2 (raw data) & $86.46 \pm 0.01 $ & $71.21 \pm 0.04$ \\
  \hline
  cld2 (preprocessed data) & $91.70 \pm 0.01$ & $77.70 \pm 0.06$ \\
  \hline
  langid (raw data) & $86.72 \pm 0.02$ & $64.02 \pm 0.06$ \\
  \hline
  langid (preprocessed data) & $88.62 \pm 0.02$ & $64.55 \pm 0.06$ \\
  \hline
  prob. model \& evidence acc. \& UI & $\mathbf{98.79 \pm 0.15}$ & $\mathbf{82.67 \pm 3.54}$ \\
\end{tabu}
\vspace{2mm}
\caption {Micro- and macro-averaged F1-scores of the best performing classification method compared to Chromium Compact Language Detector 2}
\vspace{-4mm}
\label{table:cld2_langid}
\end{center}
\end{table}

To sum up, our probabilistic model with modified Kneser-Ney smoothing with the addition of user-specific data outperforms \texttt{CLD2} by $7.09$ and \texttt{langid} by $10.17$ in the micro-averaged F1-score and by $4.97$ and $18.12$ respectively in the macro-averaged one, which is considered to be a very significant improvement. These results support the statement mentioned in the Introduction that \texttt{CLD2} is not suited well for language detection on short texts and therefore confirm the initially stated need for a better algorithm. Additionally, we show that algorithms like \texttt{langid} which work considerably well across domains have certain difficulties when competing with an algorithm designed specifically for one domain only.

\subsection{Getting More Out of the Data} \label{statistics}
In this Section, the attention is drawn to the possible applications of having a reliable short text language detection algorithm. Even though language detection by itself is an interesting and challenging task, a lot more conclusions and appealing statistics can be drawn after having it already applied on real world data. Therefore, the connections of the predicted languages with the UI language and the location are investigated.

\begin{figure}[!htb]
    \centering
        \includegraphics[width=0.9\columnwidth]{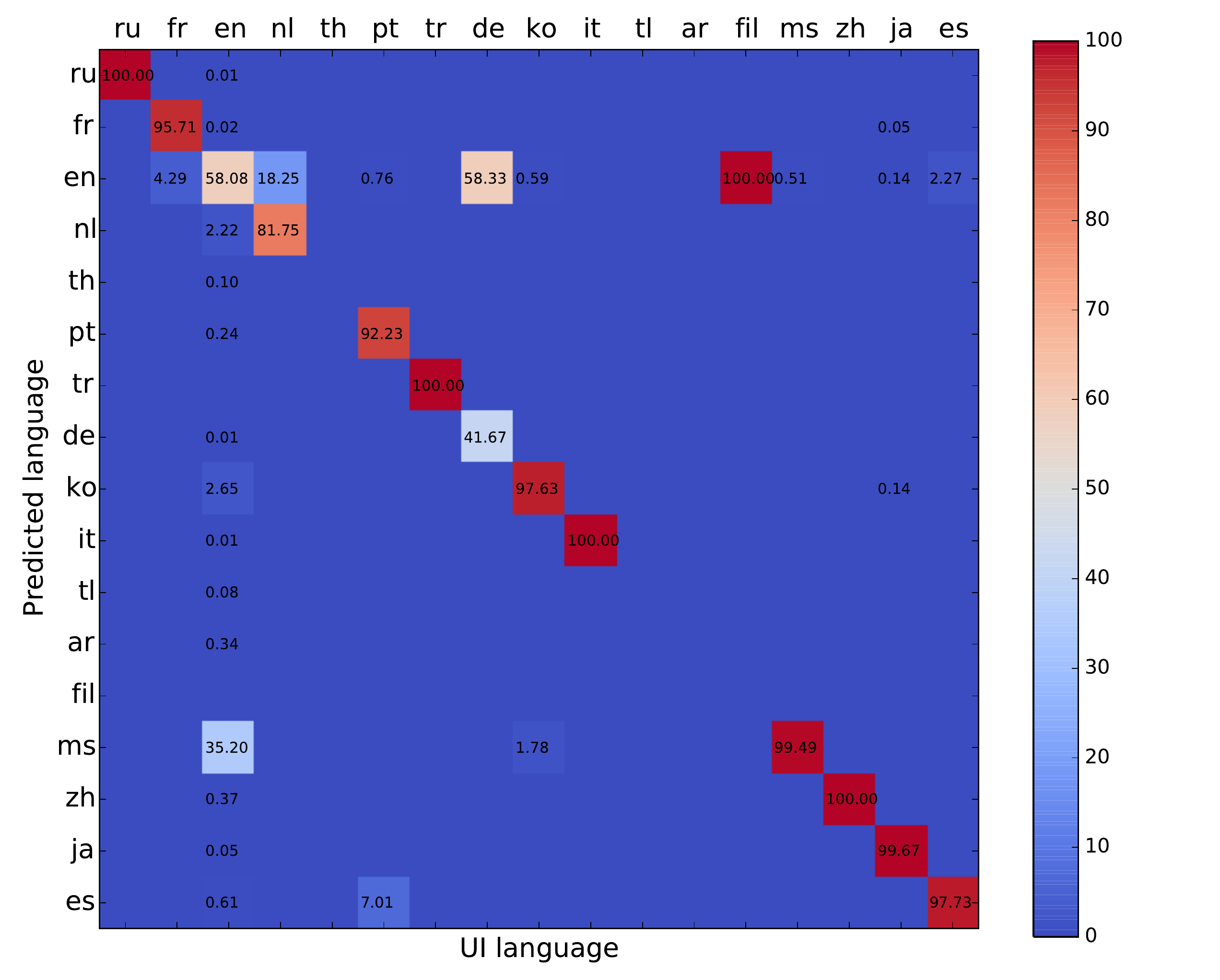}
    \caption{The percentage of predicted language vs. the UI language}
    \label{fig:lang_vs_ui}
\end{figure}

Fig. \ref{fig:lang_vs_ui} shows the relationship of the predicted language vs. the UI language. 
It is not so surprising to notice that users belonging to many different nationalities (assuming that the nationalities usually correspond the users' UI languages) tweet in English or that many users have English as their UI language, independent of which language they tweet in. On the other hand, it is interesting to see that users which tweet often in Spanish have, apart from English, Spanish and Portuguese set up as their UI languages. This illustrates the geographic and lexical similarity of those languages. It is also interesting to notice that e.g. users that tweet in Dutch have mostly English as their UI language ($78.04\%$), followed by Dutch ($21.96\%$), compared to the users who tweeted in French which mostly have French as their UI language ($93.06\%$), followed by English ($5.56\%$). This might be a good indication of how widespread the usage of English language in a particular country is.

\begin{figure}[!htb]
    \centering
        \includegraphics[width=1.15\columnwidth]{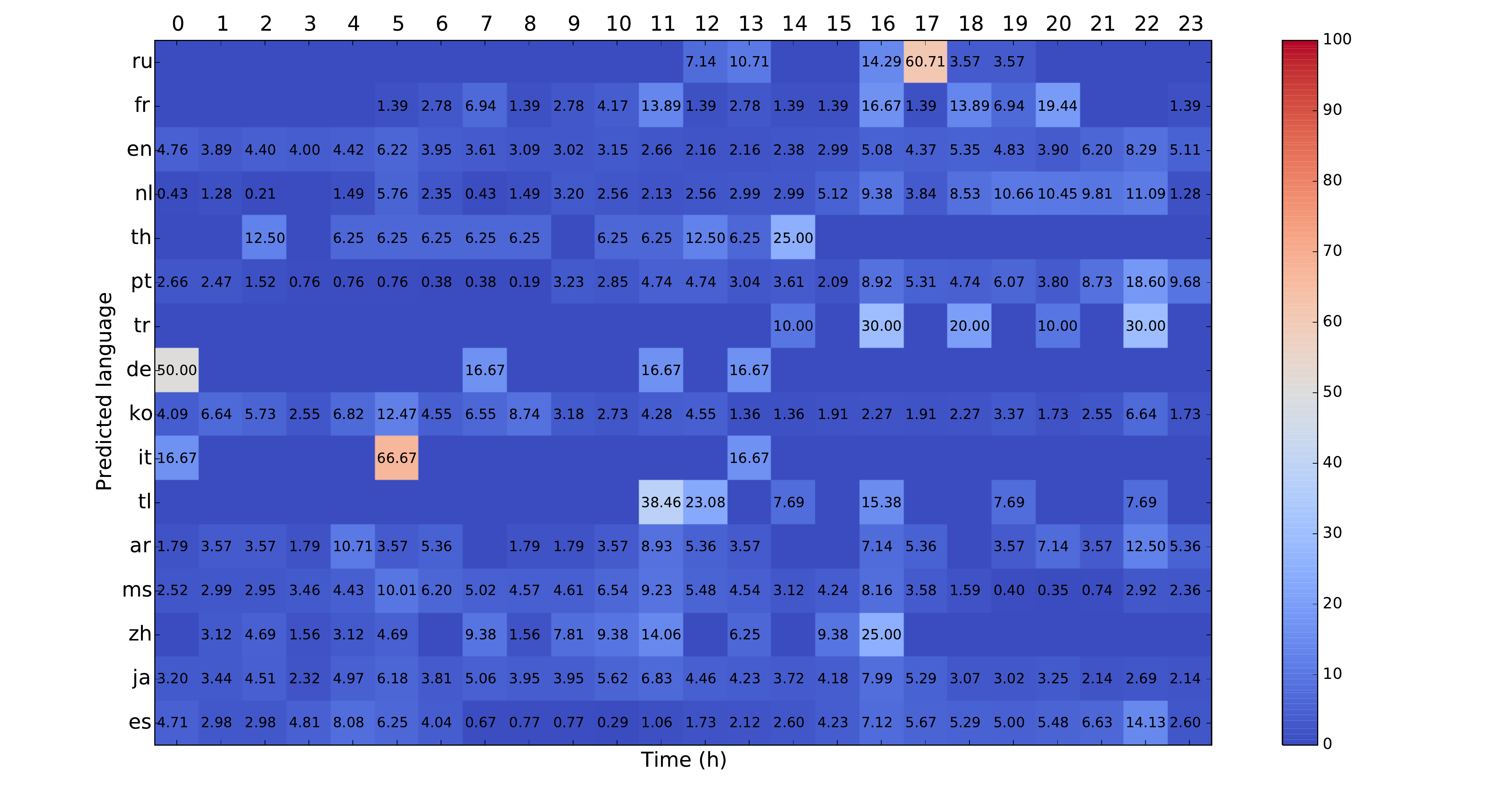}
    \caption{The percentage of predicted language vs. the time when the tweet is posted}
    \label{fig:lang_vs_time}
\end{figure}

In Fig. \ref{fig:lang_vs_time}, the predicted languages are compared with regard to the UTC time when the tweets in those languages are posted. The time is presented as hours in 0-23 range, where e.g. the number $5$ represents all tweets posted between 05:00h and 05:59h. 
It can be seen that some languages (e.g. English, Portuguese, Spanish) have rather even occurrence distribution throughout the whole day, since those languages are very widespread throughout different continents and therefore different timezones as well. On the contrary, there are no Chinese tweets after 4 pm UTC time, since then the local time in China is 12 am. As expected, with most of the languages prevalent mostly in one timezone (e.g. French, Dutch), peaks in the number of tweets can be observed in the late afternoon and evening.

\section{Conclusions}
In this paper, different algorithmic approaches to language detection for short texts in social media are investigated. The first approach includes the use of the well-known classifiers such as SVM and logistic regression and the combination of both. The second approach is based on a probabilistic model with modified Kneser-Ney smoothing, with the extension in terms of including additional information specific to a single user. The last approach is a simple dictionary based method. 
When comparing the classification performance of all the algorithms, the probabilistic model outperforms the other methods. The dictionary method achieves by far the worst results, since short tweets full of spelling mistakes, abbreviations, and acronyms do not match most of the words present in the Aspell dictionaries. The other two methods are trained directly on Twitter data, which gives them a significant advantage over the dictionary method. After introducing additional information about the users into the probabilistic model, such as the user interface language and keeping track of the languages the user previously tweeted in, the classification accuracy of the probabilistic model is increased even further. The reason why the probabilistic model with modified Kneser-Ney smoothing performs better than all the other methods presumably lies in the fact that it includes the smoothing of the n-grams probability distribution, i.e. it can effectively handle the n-grams appearing in the test data that are not present in the training data, while incorporating the relationship between lower and higher order n-grams.

The main goal of this paper was to develop a language detection algorithm aimed at short texts, since that is where most of the general language detection tools fail. Therefore, the results obtained by the above mentioned algorithms are compared with the already existing general language detection tools \texttt{Chromium Compact Language Detector 2 (CLD2)} and \texttt{langid}. Both SVM and logistic regression and especially the probabilistic model provided a substantial increase in the accuracy results compared to those two tools. This improvement becomes even more pronounced after introducing additional user-specific information into the model, which brings us one step closer to solving the task of reliably detecting the language of short texts.

\section*{Acknowledgments}
This work was supported by the Brain Korea 21 Plus Program, through the National Research Foundation of Korea funded by the Ministry of Education and by the Federal Ministry for Education and Research (BMBF) under Grant 01IS14013A-E and Grant 01GQ1115. Correspondence to KRM.

\nolinenumbers

\bibliography{bibliography}

\end{document}